\documentclass[letterpaper, 10pt, conference]{ieeeconf}

\IEEEoverridecommandlockouts                              
\usepackage[utf8]{inputenc}
\usepackage{epsfig}
\usepackage{cite}
\usepackage{subfigure}

\usepackage{amsfonts,amsthm,amsmath,amssymb,eso-pic}
\usepackage{textcomp}
\usepackage{comment}
\usepackage{graphicx}
\usepackage{gensymb}
\usepackage[font=small,skip=0pt]{caption}
\let\labelindent\relax
\usepackage[inline]{enumitem}
\usepackage{hyperref}   
\usepackage{algorithmic}
\usepackage{draftwatermark}

\graphicspath{{Figures/}}

\hypersetup{
    colorlinks=true,
    breaklinks=true,
    urlcolor=black,
    citecolor=black,
    linkcolor=blue,
    bookmarksopen=true,
    pdftitle={Neural Network and ANFIS based auto-adaptive MPC for path tracking in autonomous vehicles},
    pdfauthor={Y. Kebbati},
}

\def\BibTeX{{\rm B\kern-.05em{\sc i\kern-.025em b}\kern-.08em
    T\kern-.1667em\lower.7ex\hbox{E}\kern-.125emX}}

\newcommand{\rom}[1]{\uppercase\expandafter{\romannumeral #1\relax}}
\title{Neural Network and ANFIS based auto-adaptive MPC for path tracking in autonomous vehicles}

\author{Yassine Kebbati, Naima Ait-Oufroukh, Vincent Vigneron, Dalil Ichalal
\thanks{Yassine Kebbati, Naima Ait-Oufroukh, Vincent Vigneron, Dalil Ichalal are with IBISC-EA4526, Université Paris-Saclay, Évry, France. 
Emails: {\tt\small \{yassine.kebbati, naima.aitoufroukh, vincent.vigneron, dalil.ichalal\}@univ-evry.fr}}
}

\begin{document}
\SetWatermarkText{Preprint}
\maketitle

\begin{abstract}
Self-driving cars operate in constantly changing environments and are exposed to a variety of uncertainties and disturbances. These factors render classical controllers ineffective, especially for lateral control. Therefore, an adaptive MPC controller is designed in this paper for the path tracking task, tuned by an improved particle swarm optimization algorithm. Online parameter adaptation is performed using Neural Networks and ANFIS. The designed controller showed promising results compared to standard MPC in triple lane change and trajectory tracking scenarios.
\end{abstract}


\section{Introduction}

Autonomous driving is one of the top technologies that can radically change people's life, owing to the fact that self-driving cars can significantly reduce road accidents, alleviate traffic congestion and mitigate energy consumption and air pollution. As a consequence, researchers are racing towards building fully autonomous driving systems. One of the main components of such systems is the control module which handles lateral and longitudinal control tasks. Lateral control is a primordial part to achieve autonomous driving, which has seen extensive research over the last decades. Although there exist several simple controllers in the literature that can easily control the vehicle lateral dynamics, most of them fail to cope with external disturbances and preserve safety constraints. 

Model predictive control can handle constraints systematically setting itself as a promising tool for such tasks \cite{1,2}.  On the other hand, adaptive control has shown its ability to handle model uncertainties, disturbances and to deal with varying parameters \cite{33,3,4,5}. As the computational power increased recently, researchers aimed at integrating AI techniques to improve and adapt controllers to varying working conditions, disturbances and modeling uncertainties. These research works either use data to learn the models used in the controller design, or learn the design itself by adapting the control algorithm. For instance, Alcala \textit{et al}. \cite{6} developed an MPC controller for autonomous vehicles where they used an ANFIS approach to learn the vehicle dynamics model in the form of a polytopic representation. Similarly, authors of \cite{7} developed an adaptive MPC for lane keeping that can handle unknown steering offsets through a recursive model estimation algorithm. In the same fashion, Lin \textit{et al}. \cite{8} designed an adaptive MPC that handles changing working conditions. They used the recursive least square (RLS) algorithm to estimate the vehicle cornering stiffness and road adhesion coefficients to adapt the MPC prediction model. Brunner \textit{et al}. \cite{9} proposed a learning-based MPC for autonomous racing where racing data was used to learn a terminal cost and a safe set that guarantees recursive stability and improved performance over iterations. In a similar way, Kabzan \textit{et al}. \cite{10} proposed a learning-based MPC for autonomous racing. They used Gaussian process regression to improve the MPC prediction model using driving data after several racing laps. In \cite{11}, an MPC with low computation load has been proposed. The authors approximated the control signal with Laguerre functions and improved the tracking accuracy using exponential weight technique. Furthermore, authors of \cite{12} considered the varying road conditions and small slip-angle assumptions as a measurable disturbance in the MPC design. They used the differential evolutionary algorithm to solve the problem. 

Nonetheless, very few research papers worked on learning the control algorithm and most of them only consider constant longitudinal velocities. This paper proposes a new approach to learn and adapt the MPC algorithm to external disturbances and varying working conditions. Particularly, an improved PSO algorithm developed in \cite{19} is used to tune and optimize several parameters of the MPC algorithm to ensure optimal control performance. The resulting optimal parameters are learned using neural networks and ANFIS systems for online control adaptation. Section \rom{2} of this papers deals with vehicle modeling and MPC design, and section \rom{3} exposes the optimization of MPC parameters. Section \rom{4} presents the controller adaptation using neural networks and ANFIS approach. The proposed controller is evaluated and the results are reported and analysed in section \rom{5}. Conclusions and perspectives for future work are given in section \rom{6}.

\section{Vehicle Modeling and Controller Design}

\subsection{Vehicle Lateral Dynamics Model}

\begin{figure}[b]
\centering
\includegraphics[width=0.4\textwidth]{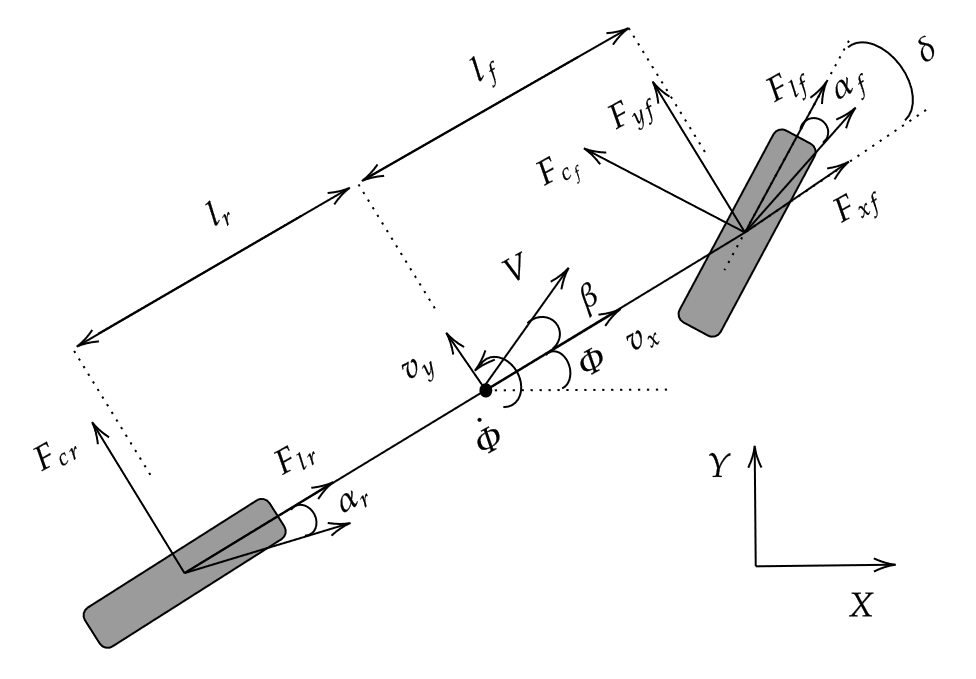}
\caption{2-DOF Bicycle dynamic model.}
\label{fig:1}
\end{figure}

Vehicle lateral control deals with the translation along the y-axis and the angular motion around the z-axis. The single track dynamic bicycle model (see Fig. \ref{fig:1}) is used for the design of the MPC controller, this model is governed by the following equations:
\begin{equation}
\label{eq1}
\left\{
    \begin{array}{ll}
 m(\Ddot{y}+\Dot{x}\Dot{\phi}) &= 2F_{yf} + 2F_{yr}\\
 I_z \Ddot{\phi} &= 2l_f F_{yf} - 2l_r F_{yr}
    \end{array}
\right.
\end{equation}

\noindent 
where, $m$ is the vehicle mass, $x$, $y$ and $\phi$ represent the longitudinal position, lateral position and the vehicle heading angle. The longitudinal forces on the rear and front wheels are $F_{yr}$ and $F_{yf}$ respectively, $l_r \backslash l_f$ are the distances between the rear$\backslash$front wheel axles and the vehicle's gravity center (CG). $F_y$ results from the nonlinear tire-road interactions, which can be linearized under the assumption of small slip angles (typically no more than 5°). The linear tire model is given as follows:
\begin{equation}
\label{eq2}
\left\{
    \begin{array}{ll}
 F_{yf} &= C_{yf} \alpha_f\\
 F_{yr} &= C_{yr} \alpha_r
    \end{array}
\right.
\end{equation}
with $C_{yf} \backslash C_{yr}$ being the stiffness coefficients for front$\backslash$rear wheels respectively, $a_{f} \backslash a_{r}$ are the respective lateral slip angles for the front$\backslash$rear wheels which are given by: 
\begin{equation}
\label{eq3}
\left\{
    \begin{array}{ll}
 \alpha_f &= \delta_f - \gamma_f\\
 \alpha_r &= \delta_r - \gamma_r
    \end{array}
\right.
\end{equation}
where $\delta_r \backslash \delta_f$ are the steering angles for the rear$\backslash$front wheels, the rear wheel is not steerable $(\delta_r = 0)$. The angles between the direction of the wheels and the longitudinal velocity are given as:
\begin{equation}
\label{eq4}
\left\{
    \begin{array}{ll}
 \tan (\gamma_f) &= \frac{\Dot{y} + l_f \Dot{\phi}}{\Dot{x}}\\
 \tan (\gamma_r) &= \frac{\Dot{y} - l_r \Dot{\phi}}{\Dot{x}}
    \end{array}
\right.
\end{equation}
Equation (\ref{eq5}) transforms the lateral velocity from the body frame to the inertial frame:
\begin{equation}
    \label{eq5}
    \dot{Y} = \dot{x} \sin{\phi} + \dot{y} \cos{\phi}
\end{equation}
The bicycle dynamic model is then obtained by replacing the above mentioned equations in (\ref{eq1}):
\begin{equation}
\label{eq6}
\left\{
    \begin{array}{ll}
        m(\Ddot{y}+\Dot{x}\Dot{\phi}) &= 2[C_{yf}(\delta_f-\frac{\Dot{y} + l_f \Dot{\phi}}{\Dot{x}}) + C_{yr}\frac{l_r \Dot{\phi}- \Dot{y}}{\Dot{x}}]\\
        I_z \Ddot{\phi} &= 2[l_{f} C_{yf}(\delta_f - \frac{\Dot{y} + l_f \Dot{\phi}}{\Dot{x}}) - l_r C_{yr} \frac{l_r \Dot{\phi} - \Dot{y}}{\Dot{x}}]
    \end{array}
\right.
\end{equation}
The corresponding state space representation is as follows: 
\begin{equation}
\label{eq7}
\left\{
    \begin{array}{ll}
        \Dot{x}_c &= A_cx_c + B_cu\\
        y_c &= C_cx_c
    \end{array}
\right.
\end{equation}
with the state vector $x_c = [\Dot{y}\ \phi\ \Dot{\phi}\ y]^T$, the output $y_c = y$ and the input $u = \delta_f$. The state, control and output matrices are given by:\\\vspace{0.06cm}
\noindent 
$A_c = \left[{\begin{array}{cccc}-2\frac{C_{yf} + C_{yr}}{m \dot{x}} & 0 & -\dot{x} - 2\frac{C_{yf} l_f -C_{yr} l_r}{m \dot{x}} & 0\\ 0 & 0 & 1 & 0\\-2 \frac{C_{yf} l_f - C_{yr} l_r}{I_z \dot{x}} & 0 & -2\frac{C_{yf} l_f^2+ C_{yr} l_r^2}{I_z \dot{x}} & 0\\1 & \dot{x} & 0 & 0\end{array}}\right],$\\\vspace{0.05cm}
$B_c = \left[\begin{array}{c} 2\frac{C_{yf}}{m} \\ 0 \\ 2\frac{C_{yf} l_f}{I_z}\\ 0 \end{array}\right],$ \hspace{5mm}
$C_c = \left[\begin{array}{cccc} 0 &  0 & 0 & 1\end{array}\right].$

\subsection{MPC Design}\label{sec:mpc}

The linear MPC, having a low computational burden, is considered in this paper. The principle of MPC is based on using the plant model to predict its response over a prediction horizon $N_p$, and generating a control sequence over a control horizon $N_c$. The control sequence, which is optimal and minimizes the tracking error, is obtained by solving a constrained convex optimization problem. Although the control sequence is optimized all along the control horizon $N_c$, only the first term is applied. The discretized form of model (\ref{eq7}) is used as the prediction model of the MPC:
\begin{equation}
\label{eq8}
\left\{
    \begin{array}{ll}
        x_{(k+1)} &= A_kx_{(k)} + Bu_{(k)}\\
        y_{(k)} &= Cx_{(k)}
    \end{array}
\right.
\end{equation}

However, the state matrix $A_k$ is updated at each iteration with the actual longitudinal velocity unlike most studies in the literature \cite{7,8,10,11}. This results in an adaptive prediction model that works for varying velocity profiles and not only constant velocities. The model is changed by adding an integrator and the output $y_{(k)}$ to the state vector to obtain an augmented model where the input becomes $\Delta u_k$:
\begin{equation}
\label{eq9}
\left\{
    \begin{array}{ll}
        \Tilde{x}_{(k+1)} &= \Tilde{A}_k  \Tilde{x}_{(k)} + \Tilde{B} \Delta \Tilde{u}_{(k)}\\
       \Tilde{y}_{(k)} &= \Tilde{C} \Tilde{x}_{(k)}
    \end{array}
\right.
\end{equation}
The new state is $\Tilde{x} = \left[\Delta x_{(k)}\  y_{(k)}\right]^T$, the new input is $\Delta \Tilde{u}_{(k)}$ and the augmented system matrices become as the following:\vspace{0.06cm}

$\Tilde{A}=
\left[\begin{array}{cc} A_k & o_m^T \\CA_k & I_{q \times q}  \end{array}\right]$,
$\Tilde{B} = \left[\begin{array}{c} B \\ CB\end{array}\right]$,
$\Tilde{C}=\left[o_m\ I_{q\times q}\right].$
The term $o_m =\overbrace{[0\ 0...0]}^\text{$n$}$ is a vector of zeros and $I_{q \times q}$ is an identity matrix where $n$, $m$ and $q$ are the number of states, inputs and outputs respectively. The control sequence over the control horizon $N_c$ is given by:
$$\Delta U = [\Delta u(k_i), \Delta u(k_i +1),..., \Delta u(k_i + N_c -1)]$$
Model (\ref{eq9}) predicts the plant behaviour over the prediction horizon $N_p$ through the following equation:
\begin{equation}
\label{eq10}
Y = F x_{(k_i)} + \Phi \Delta U
\end{equation}
where matrices $F$ and $\Phi$ are given by:
$$F = \left[\begin{array}{c} CA \\CA^2\\CA^3\\\vdots \\CA^{N_p}  \end{array}\right]$$ 
$$\Phi = \left[\begin{array}{cccc} CB&0&\hdots&0 \\CAB&CB&\hdots&0\\CA^2B&CAB&\hdots&0\\ \vdots&\vdots&\vdots&\vdots \\CA^{N_p-1}B&CA^{N_p-2}B&\hdots& CA^{N_p-N_c}B \end{array}\right]$$
The MPC problem is then defined as the following constrained optimization: 
\begin{align}
\label{eq11}
min\ &J = (R_s - Y)^T Q (R_s - Y) + \Delta U^T R \Delta U\\
\label{eq12}
s.t:\ & x(k+1) = A x(k) + B \Delta u(k) \\
\label{eq13}
    & \Delta u_{min}\leq \Delta u \leq \Delta u_{max}\\
\label{eq14}
    & u_{min}\leq u \leq u_{max}\\
\label{eq15}
    & y_{min}\leq y \leq y_{max}
\end{align}
where $Q$ and $R$ are weighting matrices, $R_s$ , $Y$ and $\Delta U$ are the reference trajectory, the output vector and the control sequence respectively. The constraints are given in terms of $\Delta U$ as the following:
\begin{equation}
\label{eq16}
M \Delta U \leq \gamma
\end{equation}
$M$ is a combination of reformulation sub-matrices and $\gamma$ groups the upper and lower bounds of the constraints. Thus, the MPC is formulated as the following quadratic programming (QP) problem:
\begin{equation}
\label{eq17}
\left\{
    \begin{array}{ll}
        &J  = \frac{1}{2}x^TEx +x^TK \\
         &Mx \leqslant \gamma
    \end{array}
\right.
\end{equation}
where $x$ is the decision variable ($\Delta u$), $E, K$ and $M$  are compatible matrices and vectors with $E$ being symmetric and positive definite. The solution of the given QP problem is based on the Hildreth's QP method \cite{13}.

\section{Controller Optimization With Improved PSO}

To optimize and tune the MPC parameters, we use an improved version of the PSO algorithm, which is a well known evolutionary algorithm in the literature \cite{14,15}. The standard algorithm is defined by:
\begin{equation}
\label{eq18}
\left\{
    \begin{array}{ll}
 v_i(k+1) =& \omega v_i(k) + c_1 r_1 (Pb_i(k)-x_i(k)) \\
 &+ c_2 r_2(Gb(k)-x_i(k))\\
 x_i(k+1)  =& x_i(k) + v_i(k+1)
    \end{array}
\right.
\end{equation}
where $v_i$ and $p_i$ are the velocity and position of particle $(i)$, which represents a solution to the optimization problem. $\omega$, $c_1$ and $c_2$ are the inertia weight, the cognitive and the social accelerations respectively, and $r_{1,2} \in [0,1]$ are random constants. $Pb$ and $Gb$ represent the best local and global positions respectively. $\omega$ and $c_{1,2}$ are constants in the classic algorithm, however, in the improved version of this work they are given by the following equations \cite{19}: 
\begin{equation}
\label{eq19}
\omega = \omega_{min} + \frac{\exp{(\omega_{max}-\lambda_1(\omega_{max}+\omega_{min})\frac{g}{G})}}{\lambda_2}
\end{equation}

\begin{equation}
\label{eq20}
\left\{
    \begin{array}{ll}
 &c_1(k+1) = c_1(k) + \alpha \\
 &c_2(k+1) = c_2(k) + \beta \\
 &\alpha = -\beta = 0.05\quad \text{for} \quad \frac gG \leq 20\%\\
 &\alpha = -\beta = 0.02\quad \text{for} \quad 20\% \leq \frac gG \leq 35\%\\
  &\alpha = -\beta = -0.035\quad \text{for} \quad 35\% \leq \frac gG \leq 75\%\\
 &\alpha = -\beta = -0.0015\quad \text{for} \quad \frac gG \geq 75\%
    \end{array}
\right.
\end{equation}

\begin{figure}[b]
\centering
\includegraphics[width=0.45\textwidth]{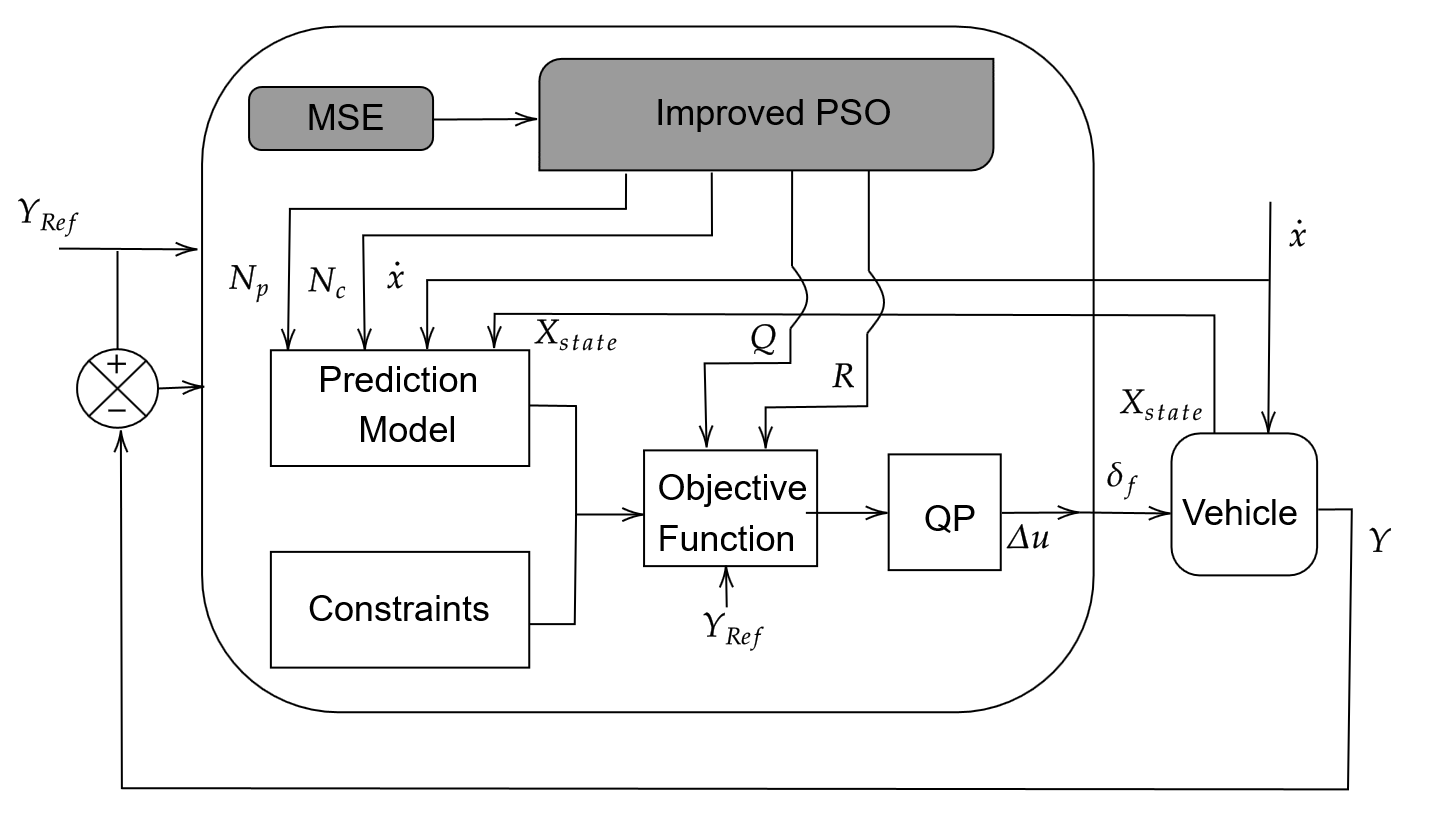}
\caption{Schematic diagram  of the optimization approach.}
\label{fig:2}
\end{figure}

\noindent where $g$ and $G$ are the current generation and the maximum generation, $\omega_{min}$ and $\omega_{max}$ are minimum and maximum inertia weights and $\lambda_{1,2}$ are adjustable constants to ensure an exponential decrease from $\omega_{max}$ to $\omega_{min}$. The advantage of this improved version over the standard one is that it enhances the overall search capabilities of the PSO algorithm \cite{19}, where the exponential decrease of $\omega$ accelerates the convergence towards the global best solution. On the other hand, increasing the cognitive acceleration $c_1$ enhances the exploration phase where particles tend to converge towards $Pb$ and increasing $c_2$ enhances the exploitation phase where particles converge towards $Gb$ and vice versa. The improved PSO algorithm is used to tune and optimize the MPC parameters which are $N_p$, $N_c$, $Q$ and $R$. Fig. \ref{fig:2} shows the optimization approach where the MSE is used as the fitness function. $Y_{ref}$ is the reference path, $Y$ is the vehicle lateral position, $\dot{x}$ is the longitudinal velocity and $X_{state}$ is the state vector. The PSO hyper-parameters used for the optimization are given in table (\ref{tab:2}), these parameters have been selected after evaluating the algorithm on typical benchmark problems like the sphere equation. To cover the majority of possible situations, the optimization is done for a variety of longitudinal speeds $\dot{x}(m/s) \in [3,27]$ and lateral references $y_{ref}(m)\in [-15,15]$. In addition, to account for external disturbances and consider different road conditions, the road adhesion coefficient is varied $\mu \in [0.5,0.9]$, and the lateral wind is also included $v_w (m/s) \in [-30,30]$. The result is a small data-set with optimal MPC parameters.    
\begin{table}[htb]
\caption{PSO hyper-parameters.} 
\label{tab:2}
\centering
\begin{tabular}{c c c} 
\hline
Parameter & Interpretation & Value\\[0.8ex] 
\hline
$N$ & Number of generations & $15$ \\[0.8ex] 
$N_{Pop}$ &  Number of particles & $20$ \\[0.8ex] 
$\omega_{max}$ & Maximum inertia weight & $0.99$ \\ [0.8ex] 
$\omega_{min}$ & Minimum inertia weight & $0.1$ \\ [0.8ex] 
$c_{1i}$ & Initial cognitive acceleration coefficient & $2$ \\ [0.8ex] 
$c_{2i}$ & Initial social acceleration coefficient & $2$ \\ [0.8ex] 
$\lambda_1$ & Constant & $30$ \\ [0.8ex]
$\lambda_2$ & Constant & $3$ \\ [0.8ex]
\hline 
\end{tabular}
\end{table}

\section{MPC Parameters Adaptation}

\subsection{MPC adaptation with neural networks}

\begin{figure}[b]
\centering
\includegraphics[width=6cm,height=4cm]{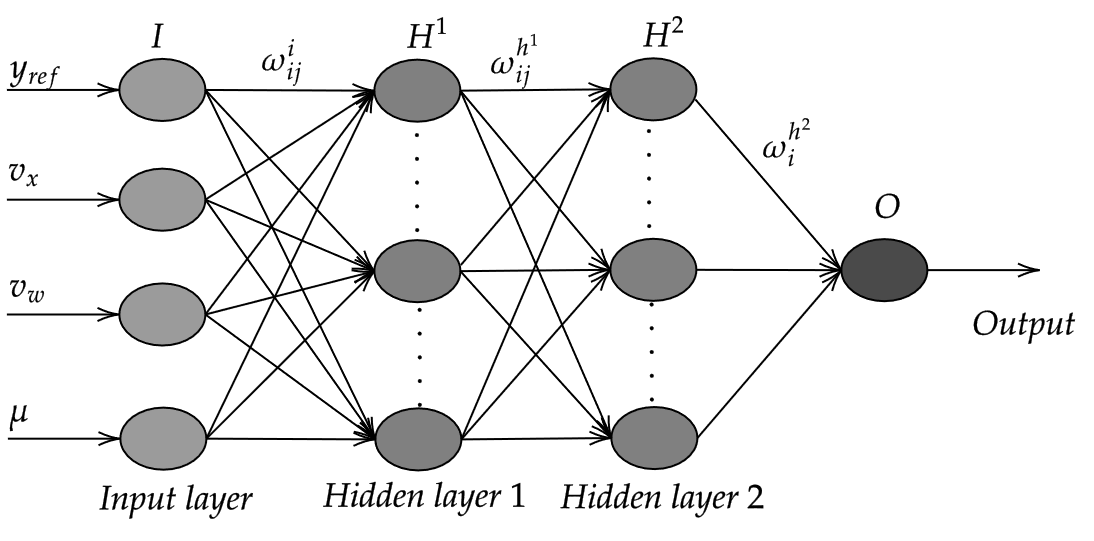}
\caption{Neural network for MPC adaptation}
\label{fig:3}
\end{figure}

After the optimisation phase which is done offline, a data-set of optimal MPC parameters is generated. In the first approach, four feedforward neural networks are used to learn the optimal MPC parameters for the sake of online adaptation. The proposed neural networks consist of four layers; the first layer contains the same four inputs, the first hidden layer consists of 16 neurons, the second hidden layer contains 8 hidden neurons and the output layer corresponds to one of the four MPC parameters: $N_p$, $N_c$, $Q$ or $R$ as shown in Fig. \ref{fig:3}. The Sigmoid activation function is used for the hidden layers and the relu function is used for the output layer since this is a regression problem for positive values only. The loss function used for the back-propagation is the MSE along with the gradient decent with momentum.

The forward pass in the neural network is governed by equations (\ref{eq21}), (\ref{eq22}) and (\ref{eq23}) respectively:
\begin{equation}
\label{eq21}
 H_j^1 = f(\sum_{i=1}^{n+1}w_{ij}^i x_i),  \; \text{for} \; j=1,..,n_{h^1}
\end{equation}
\begin{equation}
\label{eq22}
 H_j^2 = f(\sum_{i=1}^{n+1}w_{ij}^{h^1} x_i),  \; \text{for} \; j=1,..,n_{h^2}
\end{equation}
\begin{equation}
\label{eq23}
 O_j = g(\sum_{i=1}^{n+1}w_{i}^{h^2} x_i)
\end{equation}
where $f$ and $g$ are the Sigmoid and ReLU functions respectively, $\omega_{ij}$ and $x_i$ are the respective layer weights and inputs. The training is carried for 1000 epochs using a data-set of 6400 samples.

\subsection{MPC adaptation with ANFIS}

ANFIS tool is used in the second approach to learn the data-set of MPC optimal parameters in order to achieve online adaptation. Adaptive neuro-fuzzy inference systems are a hybrid AI technique that combines fuzzy logic and artificial neural networks to learn a mapping from input-output data \cite{16}. The general architecture of an ANFIS system, illustrated in Fig. \ref{fig:4}, is composed of the following:

\begin{figure}[b]
\centering
\includegraphics[width=7.5cm,height=4cm]{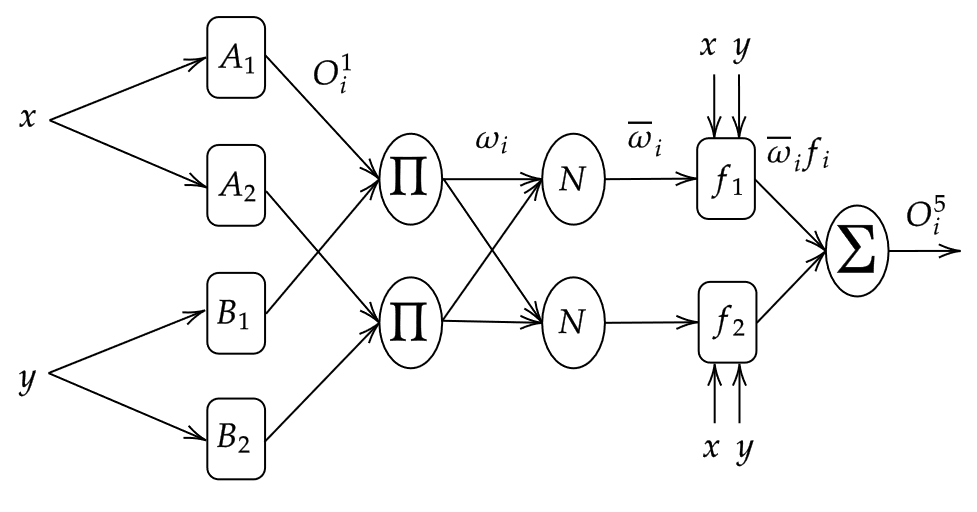}
\caption{General architecture of ANFIS}
\label{fig:4}
\end{figure}

\begin{itemize}
\item Layer 1: the fuzzification layer obtains fuzzy clusters from the input data by using membership functions as given below:
\begin{equation}
\label{eq24}
O_i^1 = \mu_{A_i}(x)
\end{equation}
with $x$ being the input to node $i$, and $A_i$ the linguistic label of the node function. $O_i^1$ is the membership function.

\item Layer 2: the rule layer computes the strengths by multiplying the membership values coming from the fuzzification layer:
\begin{equation}
\label{eq25}
\omega_i = \mu_{A_i}(x) \times \mu_{B_i}(y)
\end{equation}

\item Layer 3: the normalization layer computes the normalized strengths that belong to each rule by applying the following:
\begin{equation}
\label{eq26}
\Bar{\omega_i} = \frac{\omega_i}{\sum_i^n \omega_i}
\end{equation}
where $\Bar{\omega_i}$ is the normalized strength and $n$ is the number of nodes.

\item Layer 4: the diffuzification layer calculates the weighted values of rules through first order polynomials :
\begin{equation}
\label{eq27}
\Bar{\omega_i}f_i = \Bar{\omega_i}(p_ix+q_iy+r_i)
\end{equation}
with $f_i$ being the polynomial composed of the parameter set $\{p_i,q_i,r_i\}$ called consequence parameters.

\item Layer 5: the summation layer sums all the outputs of the diffuzification layer to obtain the ANFIS output:
\begin{equation}
\label{eq27}
O_i^5 = \frac{\sum_i \omega_i f_i}{\sum_i \omega_i}
\end{equation}
\end{itemize}

\begin{figure}[t]
\centering
\includegraphics[width=0.4\textwidth]{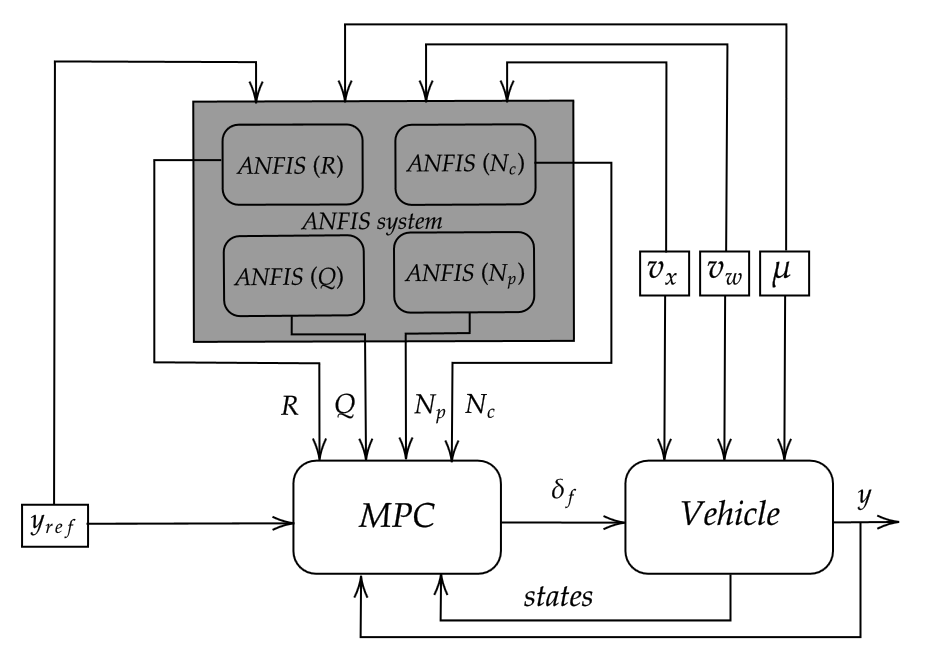}
\caption{ANFIS based adaptation approach}
\label{fig:5}
\end{figure}

The adaptation system consists of four ANFIS subsystems with four inputs each and one output corresponding to $N_p$, $N_c$, $Q$ or $R$. The hybrid training approach combining gradient decent with RLS is used to train the ANFIS and avoid local minima. The scattering method is used for input space partitioning since there are four inputs. Fig. \ref{fig:5} illustrates the proposed approach where $v_x,\ v_w,\ \mu$ and $y_{ref}$ are the longitudinal velocity, the lateral wind velocity, the road adhesion coefficient and the reference lateral position. 

\section{Results and Discussion}

For evaluating the proposed adaptive MPC controllers, a high fidelity vehicle model has been built using Vehicle Dynamics Blockset of \texttt{MATLAB}. The model consists of the 3-DOF dual track lateral dynamics block with nonlinear Pacejka tire formula \cite{18}. In addition, a simplified powertrain block is added to accommodate variable velocities through the predictive longitudinal driver block. The parameters of model (\ref{eq7}) and MPC controller are given in table (\ref{tab:2}).

\begin{table}[b]
\caption{MPC and model parameters} 
\label{tab:2}
\centering
\begin{tabular}{c c c c} 
\hline
Model parameter & Value & MPC parameter & Value\\[0.8ex] 
\hline
$m$ & 1575 (kg) &$N_p$ & $35$  \\[0.8ex] 
$I_z$ & $2875 (kg.m^2)$ &$N_c$ & $8$  \\[0.8ex] 
$l_f$ & $1.2 (m)$ & $\Delta u_{max/min}$  & $\pm \frac{\pi}{12}(rad)$ \\ [0.8ex] 
$l_r$ & $1.6 (m)$ & $u_{max/min}$ & $\pm \frac{\pi}{6}(rad)$ \\ [0.8ex] 
$C_f$ & $19000 (N/rad)$ & $R$ & $0.01$ \\ [0.8ex] 
$C_r$ & $33000 (N/rad)$ & $Q_y$ & $10$  \\ [0.8ex] 
\hline 
\end{tabular}
\end{table}

\begin{figure}[ht]
\centering
\includegraphics[width=7.7cm,height=3.5cm]{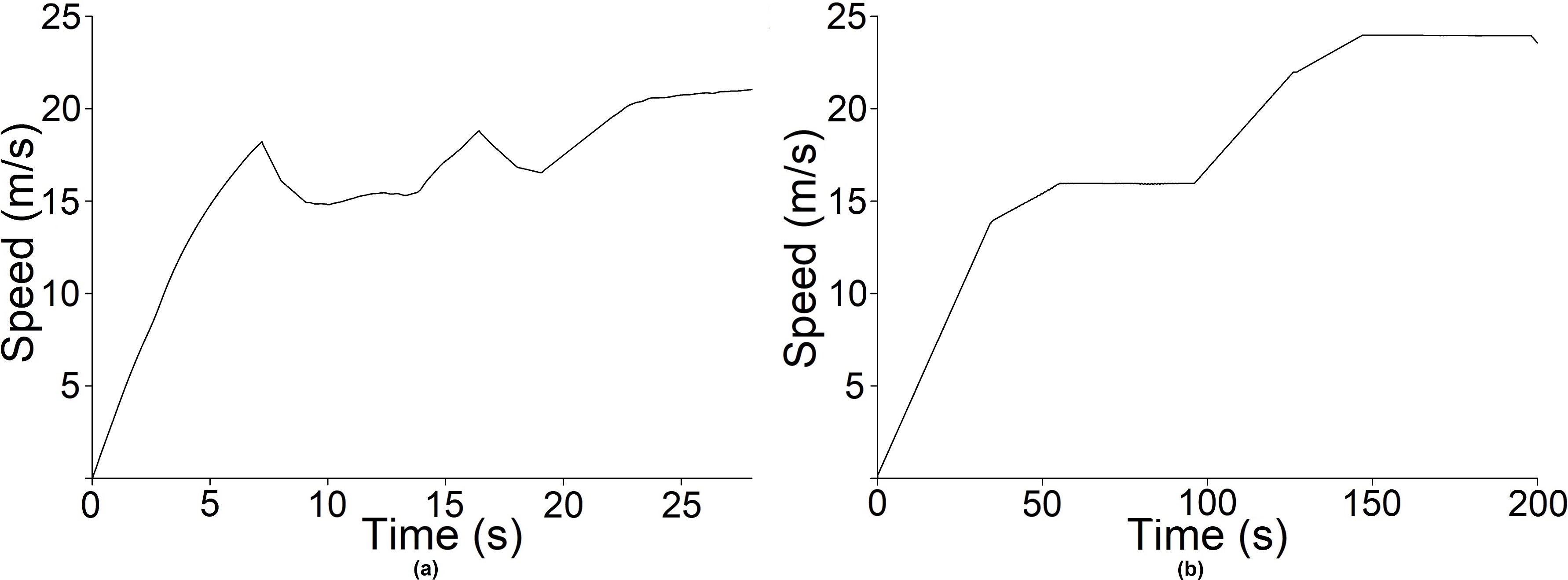}
\caption{Longitudinal velocity profile (a) : Triple lane change,\\ 
(b) : General trajectory.}
\label{fig:6}
\end{figure}

\begin{figure}[ht]
\centering
\includegraphics[width=8cm,height=7cm]{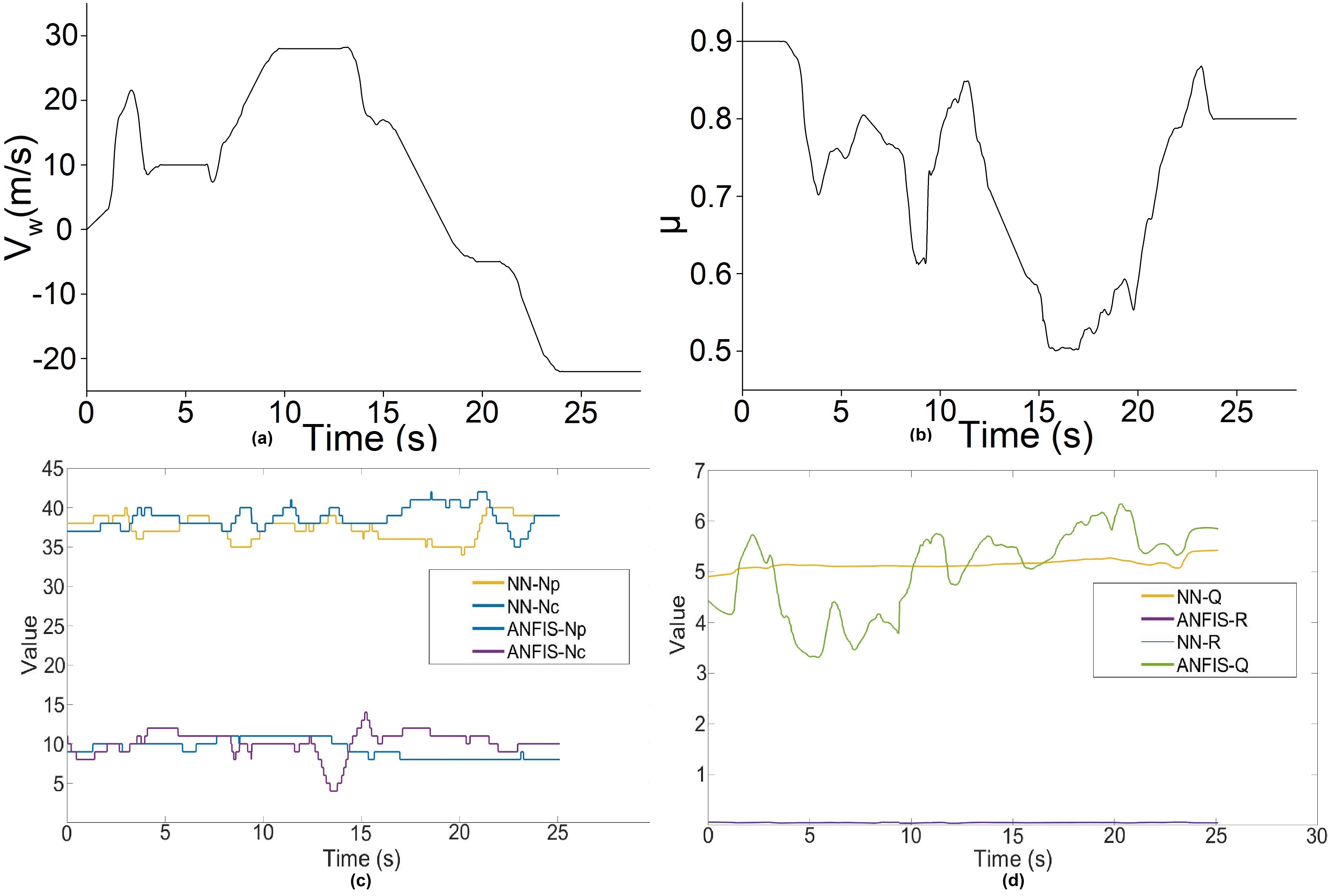}
\caption{(a) : Lateral gust profile, (b) : Road adhesion coefficient,\\ (c) : $N_c/N_p$ adaptation, (d) : $Q/R$ adaptation.}
\label{fig:7}
\end{figure}

\begin{figure}[ht]
\centering
\includegraphics[width=8cm,height=7cm]{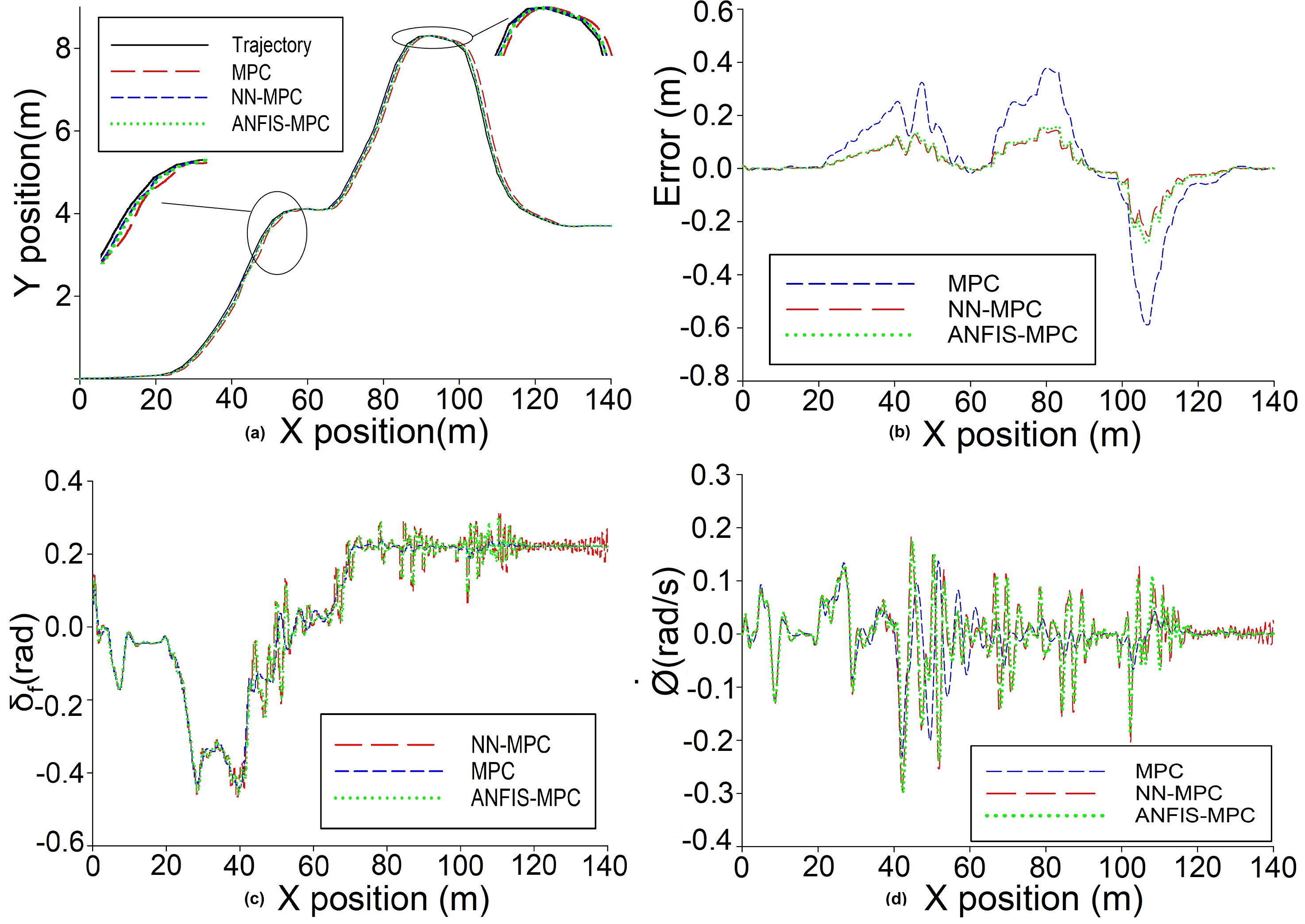}
\caption{Triple lane change ((a) : Trajectory tracking, (b) : Tracking error, (c) : Steering command, (d) : Vehicle yaw rate).}
\label{fig:8}
\end{figure}

Both NN-MPC and ANFIS-MPC are tested against the standard MPC for a triple lane change scenario which is often performed in multi-lane highways. The longitudinal velocity of the vehicle is varied according to Fig. \ref{fig:6}(a). The vehicle was exposed to wind gust and varying road adhesion coefficient, these are given in Fig. \ref{fig:7} along with the resulting adaptive Nc, Np, Q and R signals. The corresponding trajectory tracking, steering angle, tracking error and yaw rate signals are given in Fig. \ref{fig:8}. As can be seen from the figures, NN-MPC has the best tracking performance with an MSE of (0.0051) compared to (0.0062) and (0.0318) for the ANFIS-MPC and standard MPC respectively. On the other hand, ANFIS-MPC exhibits smoother control signals but is slightly less accurate. Overall, both adaptive controllers overcome the standard MPC and are much more robust and adaptive to wind disturbance and changing road conditions.

\begin{figure}[ht]
\centering
\includegraphics[width=8cm,height=7cm]{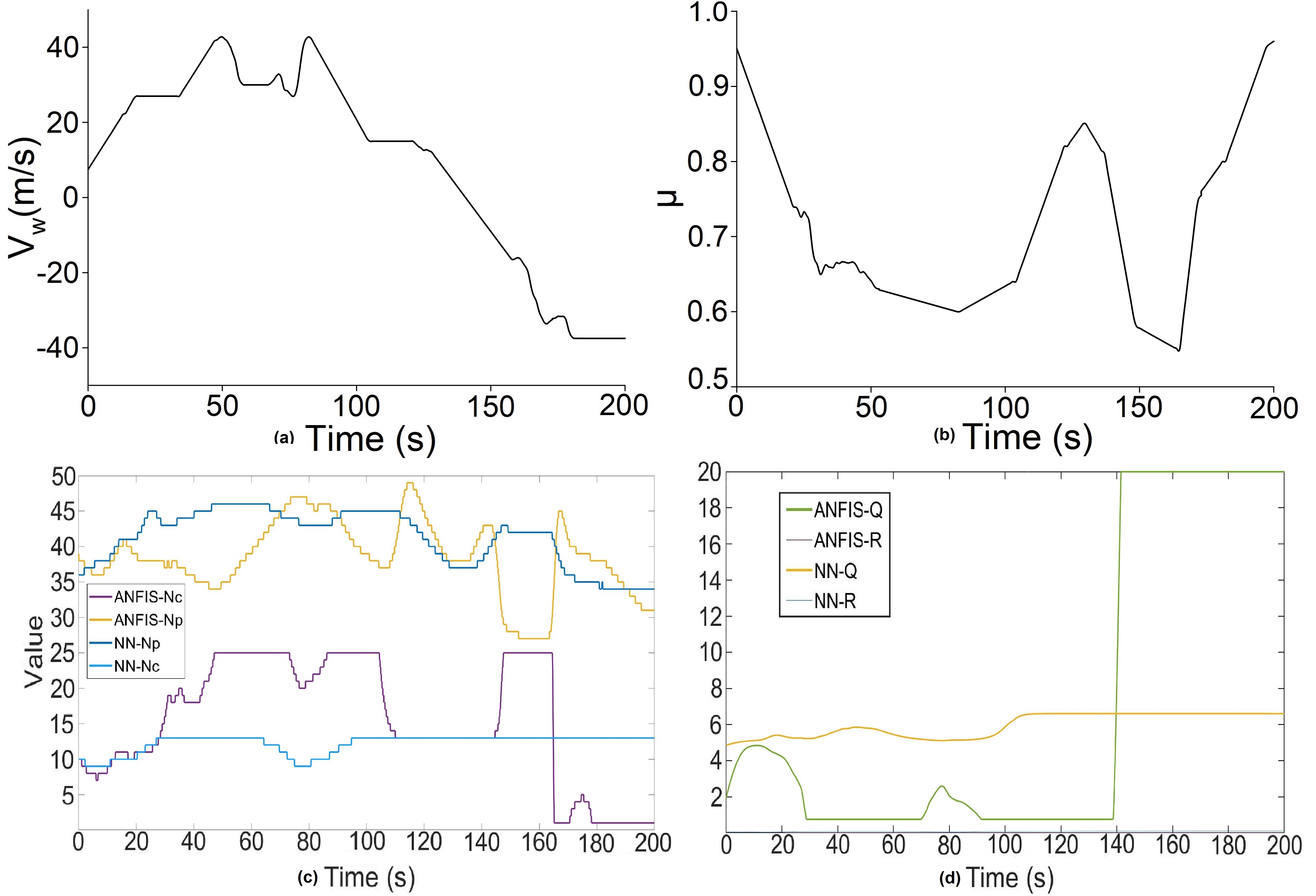}
\caption{(a) : Lateral gust profile, (b) : Road adhesion coefficient,\\ (c) : $N_c/N_p$ adaptation, (d) : $Q/R$ adaptation.}
\label{fig:9}
\end{figure}

\begin{figure}[ht]
\centering
\includegraphics[width=8cm,height=7cm]{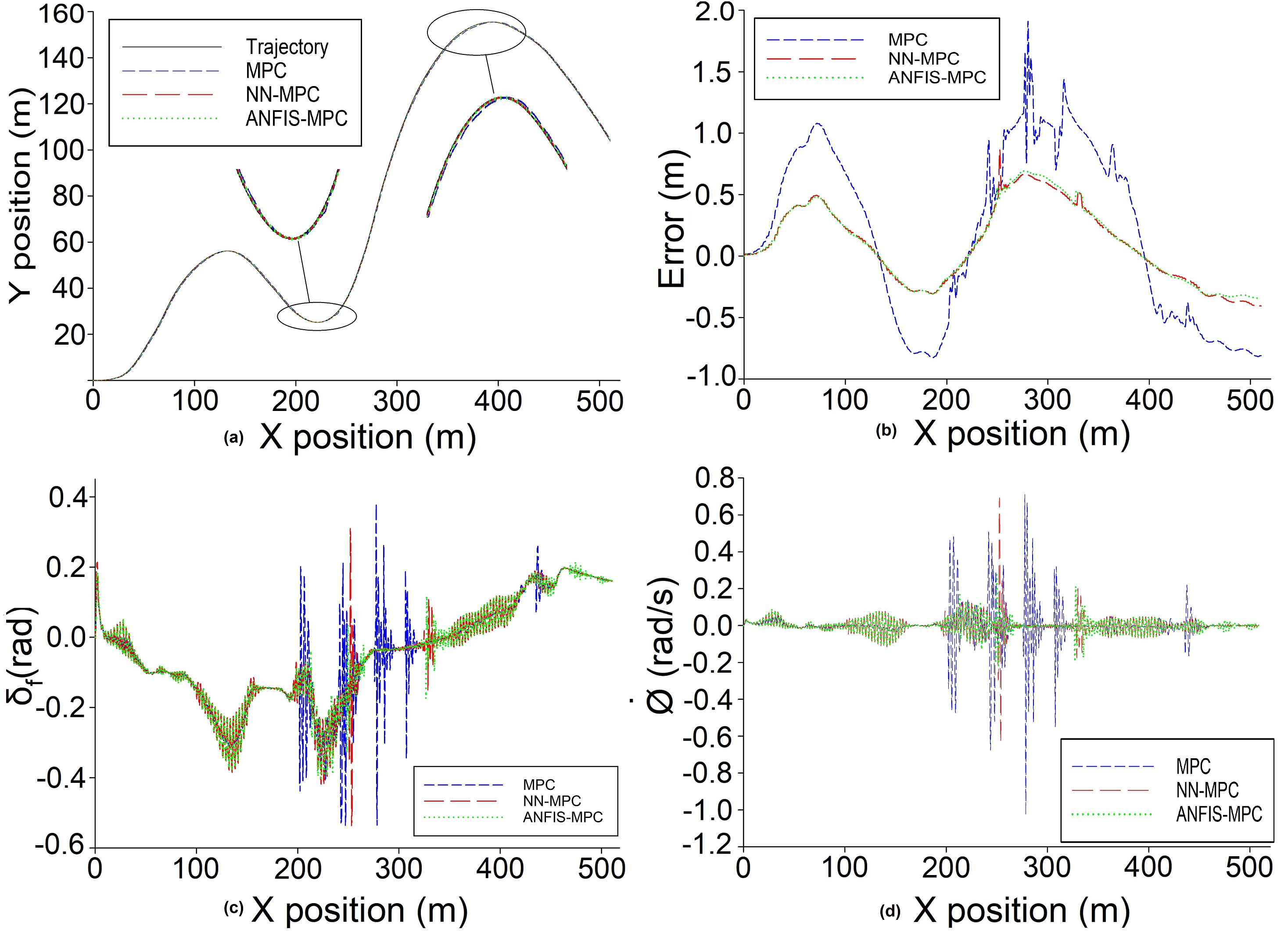}
\caption{General trajectory ((a) : Trajectory tracking, (b) : Tracking error, (c) : Steering command, (d) : Vehicle yaw rate).}
\label{fig:10}
\end{figure}

Furthermore, the controllers are tested for a general trajectory where the velocity profile is given in Fig. \ref{fig:6}(b). Wind gust, road adhesion coefficient and $Nc$, $Np$, $Q$ and $R$ signals are shown in Fig. \ref{fig:9} respectively. The corresponding results in Fig. \ref{fig:10} show similar performance to the previous test. The NN-MPC and ANFIS-MPC proved higher tracking accuracy, better disturbance rejection and are more adaptive compared to regular MPC. Similarly NN-MPC was slightly more accurate in this test with an MSE of (0.132) compared to (0.1349) and (0.5896) for ANFIS-MPC and classic MPC respectively. Contrary to NN-MPC, the main drawback of ANFIS-MPC is the curse of dimensionality. The latter is due to the exploding number of input membership functions, which makes such an approach very demanding in terms of computational power and less practical for real-time applications. 

\section{Conclusions}

In this paper, two adaptive MPC controllers are proposed for the lateral control of autonomous vehicles. An improved PSO algorithm is used to tune and optimize MPC parameters for varying working conditions and external disturbances. A data-set of optimal MPC parameters is generated and learned using two approaches; first by using MLP neural networks, and second by using ANFIS systems. The neural networks and ANFIS systems are then used for the adaptation of MPC parameters. The proposed controllers are tested against the standard MPC for a triple lane change scenario and a general trajectory. Furthermore, lateral gust is introduced with varying road conditions. The proposed controllers showed improved tracking accuracy, robustness and adaptability to disturbances and varying parameters. NN-MPC proved to be more accurate while ANFIS-MPC was found to be smooth but very demanding. Further studies shall address the improvement and learning of the MPC prediction model in the mixed longitudinal and lateral control.

\bibliographystyle{ieeeconf}
\bibliography{library}

\end{document}